\documentclass[conference]{ieeeconf}

\IEEEoverridecommandlockouts
\usepackage{titlesec}
\usepackage{tikz}
\usepackage{tikzscale}
\usepackage{tabularray}
\usepackage{amsmath}
\usepackage{amssymb}
\usepackage{float}
\usepackage{graphicx}
\usepackage{lipsum}
\usepackage{fancyhdr}
\usepackage{siunitx}
\usepackage{comment}

\title{TOPPQuad: Dynamically-Feasible Time-Optimal Path Parametrization for Quadrotors}
\author{Katherine Mao, Igor Spasojevic, M. Ani Hsieh, and Vijay Kumar %
\thanks{}
\thanks{
This works was supported by NSF Grant CCR-2112665, and the ARL DCIST CRA W911NF-17-2-0181.
Katherine Mao, Igor Spasojevic, M. Ani Hsieh, and Vijay Kumar are with the GRASP Laboratory, University of Pennsylvania, PA, 19104, USA
        {\tt\small \{maokat, igorspas, mya, kumar\}@seas.upenn.edu}}%
        }
\begin{document}


\maketitle

\begin{abstract}
Planning time-optimal trajectories for quadrotors in cluttered environments is a challenging, non-convex problem. 
This paper addresses minimizing the traversal time of a given collision-free geometric path without violating actuation bounds of the vehicle.
Previous approaches have either relied on convex relaxations that do not guarantee dynamic feasibility or have generated overly conservative time parametrizations.
We propose TOPPQuad, a time-optimal path parameterization algorithm for quadrotors which explicitly incorporates quadrotor rigid body dynamics and constraints, such as bounds on inputs (including motor thrusts) and state of the vehicle (including the pose, linear and angular velocity and acceleration).
We demonstrate the ability of the planner to generate faster trajectories that respect hardware constraints of the robot compared to planners with relaxed notions of dynamic feasibility in both simulation and hardware.
We also demonstrate how TOPPQuad can be used to plan trajectories for quadrotors that utilize bidirectional motors.
Overall, the proposed approach paves a way towards maximizing the efficacy of autonomous micro aerial vehicles while ensuring their safety.

\end{abstract}

\section{Introduction}


Autonomous micro aerial vehicles (MAVs) have demonstrated an immense potential to transform logistics in numerous domains. 
Package delivery in large metropolitan areas with dense traffic, aid distribution in disaster-stricken environments, and inventory management in large warehouses are only a few such examples. 
The unique combination of size and aerial agility makes MAVs ideal for navigating cluttered 3-D environments beyond the reach of other robots.
However, key to realizing their full potential in boosting task productivity while ensuring safe environmental interaction lies in planning missions that respect their physical limitations.
This work addresses a class of problems in planning dynamically-feasible time-optimal trajectories for MAVs.

Planning time-optimal trajectories for robots subject to actuation constraints in cluttered environments is a challenging task.
First, the presence of obstacles renders the underlying optimization problem non-convex, even for systems with linear dynamics.
Second, the high-dimensional non-Euclidean state space of most aerial vehicles makes trajectory synthesis subject to actuation constraints formidable, especially in obstacle-rich environments.
One method of addressing these hurdles is a decoupled approach to motion planning \cite{lavalle2006planning}. 
First, a sufficiently smooth collision-free geometric path is established. A corresponding optimal time parametrization respecting actuation constraints is then determined.
Our work focuses on the second stage, the minimization of the traversal time of a given geometric path while respecting individual motor thrusts bounds of the MAV.

A landmark result \cite{mellinger2011minimum} previously established the \textit{differential flatness} \cite{van1998real} of quadrotor dynamics.
This initiated a line of computationally-efficient approaches \cite{mellinger2011minimum, hehn2015real, richter2016polynomial} that optimize trajectories in the space of (piecewise) polynomials of quadrotor positions and yaw angles. 
It also demonstrated the highly non-convex dependence of motor thrusts on the underlying flat outputs, posing challenges in determining trajectories with explicit constraints on motor thrusts.
Most previous works either (a) use (convex) relaxations of actuation constraints \cite{liu2018search}, (b) use bounds on the total thrust and angular velocity instead of actual motor thrusts \cite{wang2022geometrically}, or (c) plan dynamically feasible trajectories without clear bounds on suboptimality \cite{richter2016polynomial}.
Our approach can be viewed as a generalization of \cite{richter2016polynomial} in that we use a \textit{spatially-varying} dilation of time to transform any sufficiently smooth trajectory into a feasible one.

\begin{figure}[t!]
    \centering
    \vspace{7pt}
    \includegraphics[width = \linewidth]{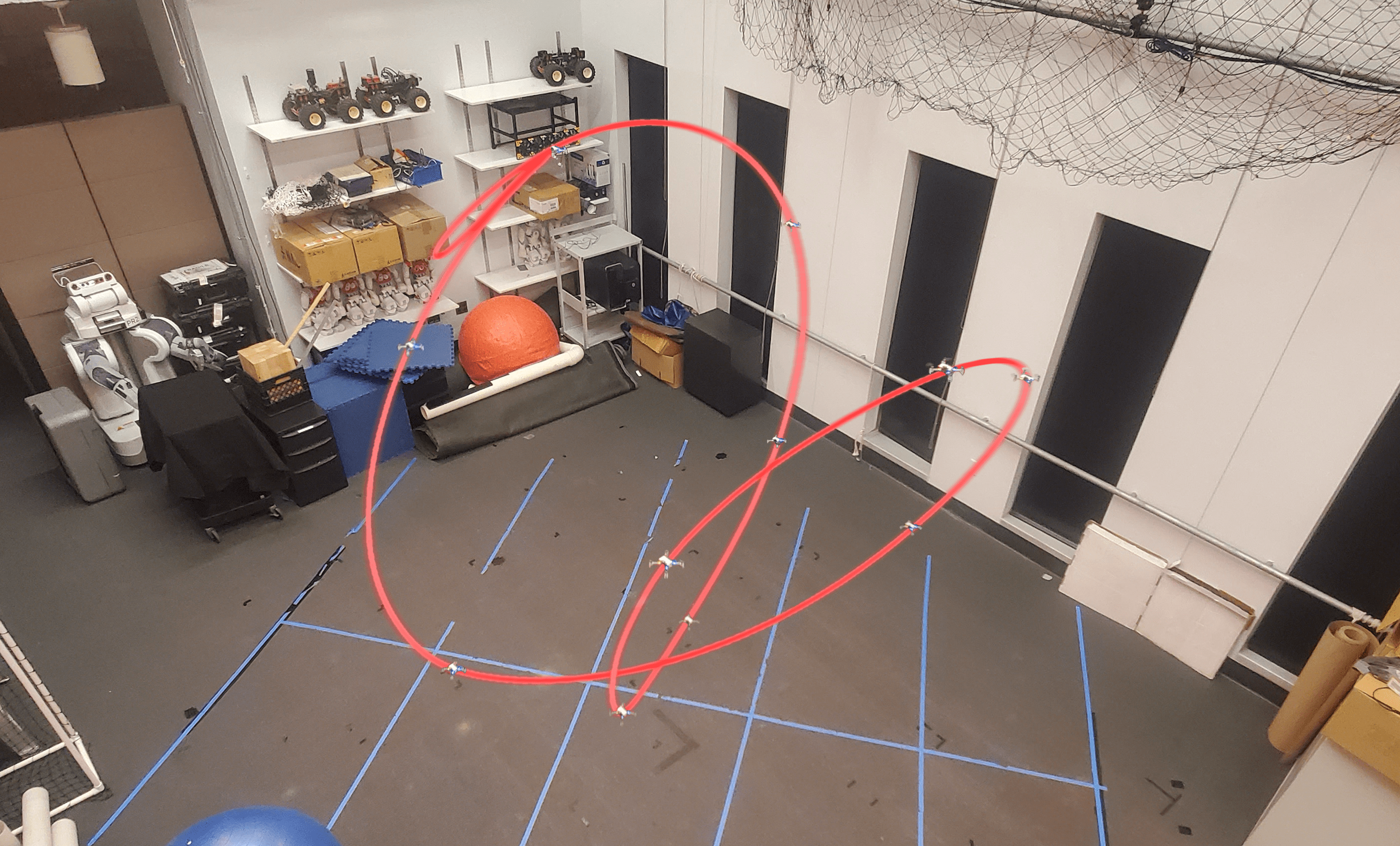}
    \caption{A time-optimal Lissajous curve trajectory computed with the TOPPQuad algorithm and tracked by a CrazyFlie quadrotor with the SE(3) Geometric Controller \cite{5717652} in a Vicon motion capture system. TOPPQuad is able to plan trajectories at speeds otherwise infeasible by conventional trajectory planners.}
    \label{fig:TOPPQUad}
\end{figure}

Optimizing spatially-varying time dilations of a trajectory is also known as the Time Optimal Path Parametrization (TOPP) problem.
Early works addressed this in the context of planning trajectories for manipulators subject to bounds on its joint velocities and torques \cite{bobrow1985time}. 
Verscheure \textit{et al} introduced the square speed profile, uncovering the hidden convexity behind such problems \cite{verscheure2009time}.
This approach has been adopted to many other robotic systems \cite{lipp2014minimum, nguyen2016time} and various computational improvements to the original method have been proposed \cite{pham2018new}.
However, none of these approaches deal with the bounds involved in planning critical rotational components of trajectories of the vehicle. This is due to the inclusion of rotations which introduces non-convex constraints, even with the aid of the square speed profile. 
Addressing the rotational aspects of trajectory generation for MAVs is one of the core themes of this paper.

Various existing works have explored planning time-optimal quadrotor trajectory with individual motor speed constraints. Ryou \textit{et al} use Gaussian Processes to learn a time-optimal obstacle-free trajectory from repeated simulation results, but any changes to the set of waypoints or obstacle space requires re-training the network \cite{ryou2021multi}. Foehn \textit{et al} formulate the planning problem by trajectory `progress'\ to find time-optimal trajectories and allows minor waypoint deviation, but in doing so assumes an sparse environment \cite{foehn2021time}. A defining characteristic of these methods is the modification of the original geometric path, a potentially problematic attribute for planning time-optimal trajectories in obstacle-dense environments.

Towards this end, we propose TOPPQuad, a TOPP algorithm that considers the full rigid body dynamics of the quadrotor.
\textbf{The contributions of this paper are as follows:}
\begin{itemize}
    \item A trajectory planner capable of refining time-optimal trajectories obtained from any arbitrary planner. 
    \item A comparison of the resulting trajectories with existing planners in simulation and real-world experimental data.
    We show that inclusion of rotational variables leads to faster dynamically feasible trajectories.
    \item {A method of planning time-optimal trajectories for a bidirectional quadrotor (i.e. one whose motor thrusts can be both parallel and antiparallel to its body $z$-axis), that seamlessly bypasses switching between flatness diffeomorphisms.}
    \item A demonstration of the trackability of our trajectories in hardware experiments
\end{itemize}

\section{Problem Formulation}


We consider the problem of minimizing the traversal time of an arbitrary sufficiently smooth geometric path for an actuation-constrained quadrotor.
A \textit{geometric} path 
\begin{equation}
\gamma : [0, S_{end}]
 \rightarrow \mathbb{R}^3
\end{equation}
is one not necessarily parametrized by time, but rather by any abstract parameter denoted by $s \in [0,S_{end}]$.
We model the quadrotor using the standard rigid body dynamics model given by 
\begin{equation}
\underbrace{
\begin{bmatrix}
\dot{\mathbf{p}} \\
\dot{\mathbf{v}} \\
\dot{\mathbf{R}} \\
\dot{\mathbf{\omega}} \\
\end{bmatrix}}_{\dot{\mathbf{x}}}
= 
\underbrace{
\begin{bmatrix}
\mathbf{v} \\
\mathbf{R}\mathbf{e}_3 \ \frac{c}{m} + \mathbf{g} \\
\mathbf{R} [\omega \times] \\
\mathbf{J}^{-1} (\tau - \omega \times \mathbf{J} \omega) \\
\end{bmatrix}}_{\mathbf{f}(\mathbf{x}, \mathbf{u})},
\label{eq:eom}
\end{equation}
where the total thrust $c \in \mathbb{R}$ and torque $\tau \in \mathbb{R}^3$ are given by 
\begin{equation}
\begin{bmatrix}
c \\
\tau \\
\end{bmatrix}
= 
\underbrace{\mathbf{F}}_{\in \mathbb{\mathbf{R}}^{4 \times 4}}
\begin{bmatrix}
u_1 \\
u_2 \\
u_3 \\
u_4
\end{bmatrix}.
\label{eq:c_and_tau}
\end{equation}
    Let $[\omega \times]$ be defined as 
\begin{equation}
   [\omega \times] = \begin{bmatrix}
0 & -w_z & w_y \\
w_z & 0 & -w_x \\
-w_y & w_x & 0 
\end{bmatrix}.
\end{equation}
In Eq. \eqref{eq:c_and_tau}, $\mathbf{F}$ is a constant \textit{scaled} control matrix, and $\mathbf{u} = [u_1, u_2, u_3, u_4]^T \in \mathbb{R}^4$ represents control inputs — the individual motor thrusts. 
The position of the vehicle's center of mass with respect to (w.r.t.) the world frame is denoted by $\mathbf{p} \in \mathbb{R}^3$, its velocity by $\mathbf{v}$, the columns of $\mathbf{R} \in SO(3)$ encode its body axes as linear combinations of world axes ($\mathbf{e}_1, \mathbf{e}_2, \mathbf{e}_3$), $\omega \in \mathbb{R}^3$ its angular velocity expressed in the \textit{body} frame of the vehicle, and $\mathbf{g}$ the gravity vector.

The minimum-time path traversal problem can then be specified as an optimization problem over the total execution time $T$ and a sufficiently smooth time parametrization function $\chi(\cdot)$: 
\begin{equation}
\begin{aligned}
& \hspace{-3mm} \min_{\substack{T > 0,\\ \chi : [0, T] \rightarrow [0, S_{end}]}}
 T  \\
& \hspace{6mm} s.t. \hspace{1mm} \chi(0) = 0, \ \chi(T) = S_{end}, \\
& \hspace{12mm} \chi(\cdot) \ \text{increasing on } [0, S_{end}], \\
& \hspace{12mm} \mathbf{p}(t) = \gamma( \chi(t) ), \quad \text{(path following constraint)} \\
& \hspace{12mm} \dot{\mathbf{x}}(t) = \mathbf{f}(\mathbf{x}(t), \mathbf{u}(t)), \quad \text{(dynamics constraint)} \\
& \hspace{12mm} \mathcal{H}(\mathbf{x}(t), \chi(t)) \leq 0, \hspace{12mm} \text{(state constraints)}\\
& \hspace{12mm} u_{min} \leq \mathbf{u}(t) \leq u_{max} \hspace{4mm} \text{(actuation constraint)}.
\end{aligned}
\label{eq:time_opt}
\end{equation}
The relation $\mathcal{H}(\mathbf{x}(t), \chi(t)) \leq 0$ collects state constraints such as bounds on the velocity, acceleration, and angular velocity of the vehicle.
Additional requirements could include pose constraints at specified points along $\gamma$ that ensure the robot can pass through narrow gaps.
\section{Methodology}

\subsection{Problem Reparametrization} \label{sec:reparametrization}

Solving Problem \eqref{eq:time_opt} for many types of robots is easier upon introduction of the \textit{square speed profile} \cite{verscheure2009time,lipp2014minimum}, defined as 
$h(s) := \left( \frac{ds}{dt}(s) \right)^2$
$\forall s \in [0, S_{end}]$.
$h$ represents the square speed of the robot as a function of distance traversed along the path. 
One motivation for the square speed profile comes from the elementary relation $\Delta v^2 = 2 a \Delta s$; a bound on the maximum acceleration is equivalent to a bound on the Lipschitz constant of $v^2$ (i.e. $h$) w.r.t. $s$ — a linear (and thus convex) constraint \cite{verscheure2009time,consolini2017optimal, AsymptoticTOPP}. 

In addition, we rewrite the orientation kinematics ($\dot{\mathbf{R}} = \mathbf{R} [\omega \times]$) as
\begin{equation} \label{eq:quaternion_tkinematics}
\dot{\mathbf{q}}(t) = \frac{1}{2} \Omega( \omega ) \mathbf{q}(t),
\end{equation}
where $\mathbf{q}(t) \in S^3$ is a quaternion representing the rotation matrix $\mathbf{R}$, and 
\begin{equation} 
\Omega( \omega ) 
:= 
\begin{bmatrix}
0 & -\omega^T \\
\omega & -[\omega \times] \\
\end{bmatrix}.
\end{equation}

Let $(\cdot)' = d(\cdot)/ds$ and $(\cdot) = d(\cdot)/dt$.
Using the relation 
\begin{equation}
\frac{d}{dt} = \frac{ds}{dt} \frac{d}{ds} = \sqrt{h(s)} \frac{d}{ds},   
\end{equation}
 we rewrite the objective and dynamics as follows. 
The objective function can be recovered as 
\begin{equation}
T = \int_{0}^{S_{end}} dt = \int_{0}^{S_{end}} \frac{dt}{ds} ds = \int_{0}^{S_{end}} \frac{1}{\sqrt{h(s)}} ds.
\end{equation} 
We next unpack and rewrite the dynamic constraints in Problem \eqref{eq:time_opt}. 
For the translational dynamics, the crucial relation is $\mathbf{p}(s) = \gamma(s)$.
Differentiating, we get
\begin{equation} \label{eq:dotp_s}
\dot{\mathbf{p}} = \sqrt{h(s)} \mathbf{p}'(s) = \sqrt{h(s)} \gamma'(s),
\end{equation}
\begin{equation}
\dot{\mathbf{v}} = \frac{1}{2} \gamma'(s) h'(s) + \gamma''(s) h(s).
\end{equation}
Substituting these into the second line of Eq. (\ref{eq:eom}), 
\begin{equation} \label{eq:translation_dynamics}
\frac{1}{2} \gamma'(s) h'(s) + \gamma''(s) h(s) = \mathbf{R}(s)\mathbf{e}_3 \ \frac{c(s)}{m} + \mathbf{g}.
\end{equation}
For the rotational dynamics, we write
\begin{equation} \label{eq:quaternion_skinematics}
\mathbf{q}'(s) = \frac{1}{2} \Omega( [\omega(s) \times] ) \mathbf{q}(s).
\end{equation}
where the relationship between $\omega$ from Eq. \eqref{eq:quaternion_tkinematics} and \eqref{eq:quaternion_skinematics} is 
\begin{equation} \label{eq:dotomega_s}
\omega(t) = \sqrt{h(s)} \omega(s).
\end{equation}
Noting the parallel between Eq. \eqref{eq:dotp_s} and \eqref{eq:dotomega_s}, and defining 
\begin{equation}
    \alpha(s) := \omega'(s),
\end{equation}
we get the relation 
\begin{equation}
\dot{\omega} = \frac{1}{2} \omega(s) h'(s) + \alpha(s) h(s).
\end{equation}
This ultimately allows us to rewrite line 4 of Eq. \eqref{eq:eom} as
\begin{equation} \label{eq:rotation_dynamics}
\tau = \mathbf{J} \underbrace{\left( \frac{1}{2} \omega(s) h'(s) + \alpha(s) h(s) \right)}_{ = \ \dot{\omega}} + \omega(s) \times \mathbf{J}\omega(s) \ h(s).
\end{equation}
Combining Eq. (\ref{eq:translation_dynamics}) and (\ref{eq:rotation_dynamics}), we get that the {individual motor thrusts at point $s$ along the path are related to the translational and angular accelerations via 
\begin{equation} \label{eq:collected_dynamics}
\begin{aligned}
\begin{bmatrix}
m(\frac{1}{2} \gamma'(s) h'(s) + \gamma''(s) h(s) - \mathbf{g}) \\
\mathbf{J} \left( \frac{1}{2} \omega(s) h'(s) + \alpha(s) h(s) \right) + \omega(s) \times \mathbf{J}\omega(s) \ h(s) \\
\end{bmatrix}
& \\
= \begin{bmatrix}
\mathbf{R}(s) \mathbf{e}_3 & 0 \\ 
0 & \mathbf{I}_3 \\
\end{bmatrix} 
\mathbf{F} \mathbf{u} \hspace{20mm} & 
\end{aligned}
\end{equation}

\subsection{Numerical Implementation}

To solve Problem \eqref{eq:time_opt}, we use the following decision variables:
\begin{equation} \label{eq:our_decision_vars}
h(\cdot), \ h'(\cdot), \ \mathbf{q}(\cdot), \ \omega(\cdot), \ \alpha(\cdot), \ \mathbf{u}(\cdot),
\end{equation}
which are all functions on $[0, S_{end}]$ as related by equations in \ref{sec:reparametrization}.
We approximately represent these \textit{functional} decision variables by their values at a finite set of $N$ grid points $D = (0 = s_0 < s_1 < \cdots < s_N = S_{end})$ spaced $\Delta s$ apart. 
Thus, $h_{i} := h(s_i)$ and likewise for the remaining variables in Eq. (\ref{eq:our_decision_vars}).
The differential constraints are approximated using forward Euler integration. 
As such, the effective dynamic constraints for $h(\cdot)$ and $\omega(\cdot)$ become 
\begin{equation} \label{eq:h_w_dynamics}
\begin{bmatrix}
h_{i+1} \\
\omega_{i+1} \\
\end{bmatrix}
=
\begin{bmatrix}
h_{i} + h_{i}^{'} \ \Delta s \\
\omega_{i} + \alpha_{i} \ \Delta s
\end{bmatrix}
\end{equation} 
\begin{equation} \label{eq:discretized_collected_dynamics}
\begin{bmatrix}
m(\frac{1}{2} \gamma_i^{'} h_{i}^{'} + \gamma_{i}^{''} h_i - \mathbf{g}) \\
J \left( \frac{1}{2} \omega_i h_i^{'} + \alpha_i h_i \right) + \omega_i \times J\omega_i \ h_i \\
\end{bmatrix}
= 
\begin{bmatrix}
\mathbf{R}_i \mathbf{e}_3 & 0 \\ 
0 & \mathbf{I}_3 \\
\end{bmatrix} 
\mathbf{F} \mathbf{u}_i.
\end{equation}
Actuation constraints amount to $\mathbf{u}_i \in [u_{min}, u_{max}]$,
and the objective function, 
the total execution time, equals 
\begin{equation}
T = \sum_{i = 0}^{N-1} \frac{2\Delta s}{\sqrt{h_i} + \sqrt{h_{i+1}}}.
\end{equation}
The critical numerical approximation is in the rotational kinematics. 
We approximate the dynamics of $q$ via 
\begin{equation} \label{eq:quaternion_discrete_dynamics}
\mathbf{q}_{i+1} = \frac{(\mathbf{I}_4 + \frac{\Delta s}{2} \Omega(\omega_i)) \ \mathbf{q}_i}{\sqrt{1 + \frac{\Delta s^2}{4} || \omega_i ||_2^2}},
\end{equation}
where $\mathbf{I}_4$ denotes the $4 \times 4$ identity matrix.
The numerator of Eq. \eqref{eq:quaternion_discrete_dynamics} performs a first-order Euler integration of quaternion kinematics, then the denominator projects this value back onto $S^3$, the manifold of quaternions.


\section{Simulation Setup}

We benchmark TOPPQuad against a family of flatness-based planners and alternative TOPP procedures based on convex relaxations of the dynamic constraints.
All planners are written in Python and computed on a laptop with an i7-6700HQ CPU. We use quadrotor parameters from the CrazyFlie 2.0 from Bitcraze \cite{8046794}, including of actuation constraints $\mathbf{u}_i = [0, 0.14375]N$. The flatness-based planners are implemented with a linear least-squares program \cite{2020SciPy-NMeth}, the TOPP methods with \texttt{cvxpy} \cite{diamond2016cvxpy}, and our TOPPQuad algorithm with the CasADi interface to IPOPT \cite{Andersson2019}. The planners are compared on a set of 200 randomized trajectories, each formed by interpolating four waypoints sampled from within a $10m \times 10m \times 10m$ box. 

\subsection{Baselines}

A defining characteristic of TOPPQuad is the generation of time-optimal trajectories that do not modify the geometric path of the original trajectory. As such, we compare only against trajectory planners that maintain this characteristic.
We select three sets of commonly-used differential flatness-based planners as baselines: minimum snap, minimum jerk, and minimum acceleration \cite{richter2016polynomial}. Within each set, we apply the following three types of constraints, while ensuring the geometric path remains fixed:

\subsubsection{An unconstrained flatness-based trajectory planner} Waypoint visit times are pre-assigned, with the time interval between consecutive waypoints determined from their Euclidean distance and a nominal velocity $v$. All geometric paths are based on this output.

\subsubsection{An upper bound on the maximum speed along the path} A convex relaxation of the TOPP problem on the path from 1) is solved by placing an equivalent bound on $h$.
\begin{equation}
\begin{aligned}
\min & \sum_{i = 0}^{N-1} \frac{2\Delta s}{\sqrt{h_i} + \sqrt{h_{i+1}}} + \lambda \sum_{i = 0}^{N} (h_{i}^{(3)})^2 \quad\\
s.t. \quad 
& 0 \leq h_{i} \leq \frac{v_{max}^2}{|| \gamma_i' ||_2^2} \quad \forall i \leq N.
\label{eq:topp_vel}
\end{aligned}
\end{equation}

\subsubsection{An upper bound on the maximum speed and total thrust} 
Additional bounds on maximum total thrust are added to the TOPP problem in 2).
\begin{equation}
\begin{aligned}
\min & \sum_{i = 0}^{N-1} \frac{2\Delta s}{\sqrt{h_i} + \sqrt{h_{i+1}}} + \lambda \sum_{i = 0}^{N} (h_{i}^{(3)})^2 \quad\\
s.t. \quad 
& 0 \leq h_{i} \leq \frac{v_{max}^2}{|| \gamma_i' ||_2^2} \\
& \left|\left| \frac{1}{2} \gamma_{i}^{'} h_{i}^{'} + \gamma_{i}^{''} h_{i} - \mathbf{g} \right|\right|_2 \leq 4 u_{max}\quad \forall i \leq N.\\
\label{eq:topp_acc}
\end{aligned}
\end{equation}

For 2) and 3), we represent the square speed profile as a third-order integrator to ensure a $\mathcal{C}^3$ trajectory and thus continuity up to jerk. We add a regularization term $\lambda \sum_{i = 0}^{N} (h_{i}^{(3)})^2$ to the objectives of Eq. \eqref{eq:topp_vel} and \eqref{eq:topp_acc} for a small positive $\lambda$ to ensure our optimization problems are numerically well conditioned. The sub-optimality of the solutions of convex relaxations decreases as $\lambda \downarrow 0$, but non-vanishing values of $\lambda$ are necessary for the solver to converge. In our experiments, we chose $\lambda = 10^{-1}$.

We refer to ``$\alpha$-scaling'' ($\alpha$) as the method introduced in \cite{richter2016polynomial}, which slows down the time uniformly across the trajectory until all motor thrusts lie within allowed ranges. 
This is equivalent to multiplying the square speed profile and its spatial derivatives by a suitable quantity (less than one). We perform $\alpha$-scaling on each of the baselines mentioned above to enforce dynamic feasibility.
\\
\subsection{Initial Guess}
TOPPQuad requires an initial guess of variables in Eq. \eqref{eq:our_decision_vars}, the quality of which can greatly affect optimizer performance. For the following comparisons, the initial guess is supplied from the baseline trajectories. To determine a `good' initial guess, we examine the effect of three parameters: the planner's nominal velocity $v$, the constraints on the planner, and $\Delta s$ determined by the number of discretization points.
Failure is considered to be one of two cases: when the CasADi solver fails to find a solution, or when the solver settles on a local minimum
with a significantly slower traversal time than the initial guess. 

We see that of three nominal velocities, $v = 1m/s$ offers the best performance in both iteration count and success rate, as shown in Table \ref{tab:consistency_table}.
A further examination shows that of the three cases considered, only at $v=1m/s$ do the motor thrusts of the initial guess consistently stay within the input bounds of the quadrotor. 

\begin{table}[h]
\centering
\begin{tabular}{clll}
\textbf{Nominal Velocity ($v$)} & \multicolumn{1}{c}{\textbf{1m/s}} & \multicolumn{1}{c}{\textbf{3m/s}} & \multicolumn{1}{c}{\textbf{5m/s}} \\ \hline
\textbf{Avg Iterations}   & 473                         & 659                          & 1138                         \\
\textbf{Success Rate}        & 0.981                                & 0.918                                 & 0.609   \\
\hline
\end{tabular}
\caption{TOPPQuad Initial Guess Success - Velocity Comparison}
\label{tab:consistency_table}
\vspace{-15pt}
\end{table}

We then compare in Table \ref{tab:optimizization_success} the algorithm's optimization success rate given initial guesses from four planners:
1) a minimum snap (MS) trajectory computed with $v = 1m/s$,
2) an $\alpha$-scaled minimum snap trajectory computed with some higher $v$,
3) an $\alpha$-scaled TOPP trajectory with only speed bounds, and 
4) an $\alpha$-scaled TOPP trajectory with speed and acceleration bounds. We see that the initial guesses from 1) have the highest success rates.

\begin{table}[h]
\centering
\begin{tabular}{cllll}
\textbf{}             & \multicolumn{1}{c}{\textbf{MS (1 m/s)}} & \multicolumn{1}{c}{\textbf{$\alpha$ MS}} & \multicolumn{1}{c}{\textbf{$\alpha$ TOPP vel}} & \textbf{$\alpha$ TOPP acc} \\ \hline
\textbf{Success Rate} & 0.991                                         & 0.981                                              & 0.936                                                & 0.891     \\             
\hline
\end{tabular}
\caption{TOPPQUad Initial Guess Success - Planner Comparison}
\label{tab:optimizization_success}
\vspace{-15pt}
\end{table}

Based on these two observations, we postulate that an initial guess with motor speeds far from the bounds of the input constraints is more likely to succeed, and in fewer iterations. We proceed using the $v = 1m/s$ flatness-based planner initial guess for the remaining experiments.


For the trajectories in this experiment, we find $N=300$ ($\Delta s 
\approx 0.025$) leads to adequate optimizer convergence. Selecting $N$ too coarsely will see IPOPT frequently fail to converge, as the discrete dynamics in Eq. \eqref{eq:discretized_collected_dynamics}, \eqref{eq:quaternion_discrete_dynamics} no longer accurately model the continuous dynamics in Eq. \eqref{eq:collected_dynamics}. On the other hand, selecting $N$ too finely will drastically increase run time for diminishing return in trajectory feasibility with a controller in the loop.

\section{Comparison of Planning Algorithms}

In this section, we first demonstrate the effectiveness of our algorithm along two metrics: dynamic feasibility and time optimality. For dynamic feasibility, we compare our optimized trajectories against the aforementioned baselines. For time optimality, we compare our algorithm against only the dynamically feasible $\alpha$-scaled baselines. Next, we extend our formulation to trajectory generation for bidirectional quadrotors. In addition, we demonstrate the flexibility of the initial guess with real-world trajectories from odometry data.

For a fair comparison, we  use $v=5m/s$ as the nominal velocity for the flatness-based planners and likewise. Thus, we show that given the same maximum velocity, our TOPPQuad trajectories are able to take fuller advantage of the quadrotor's flight envelope and achieve higher average speeds. The computed trajectories average $27.5m$ in length for the minimum snap planner, $26.8m$ for the minimum jerk planner, and $21.4m$ for the minimum acceleration planner. 

\subsection{Dynamic Feasibility}
\begin{figure}[h]
    \centering
    \includegraphics[width=\linewidth]{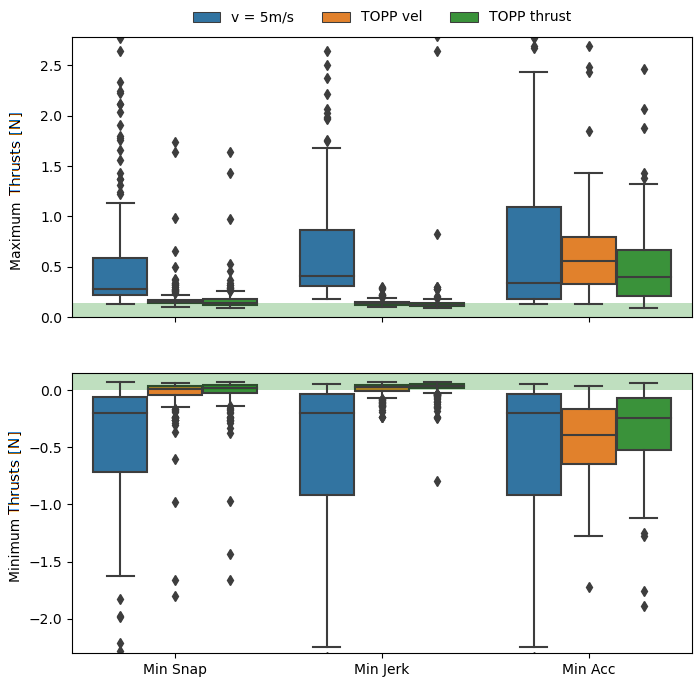}
    \caption{The distribution of maximum (\textbf{top}) and minimum (\textbf{bottom}) thrusts over 200 randomized trajectories for three sets of planners: minimum snap (\textbf{left}), minimum jerk (\textbf{middle}), and minimum acceleration (\textbf{right}). In each set, we compare a flatness-based planner at $v=5m/s$ (\textbf{blue}), a TOPP planner subject to velocity constraints (\textbf{orange}), and a TOPP planner subject to velocity and thrust constraints (\textbf{green}). No planner consistently plans trajectories that stay within motor thrust bounds.}
    \label{fig:motor_speed_infeasible}
\end{figure}

\begin{figure}[h]
    \centering
    \vspace{5pt}
    \includegraphics[width=\linewidth]{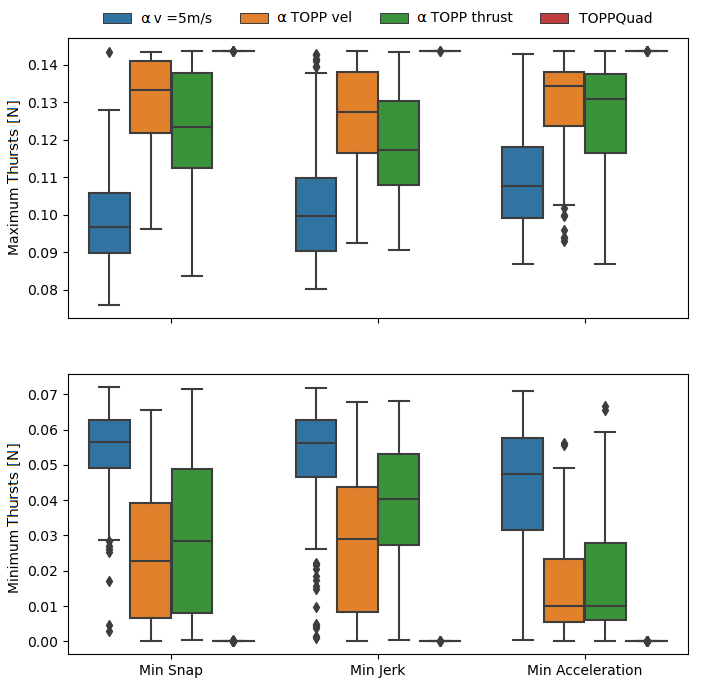}
    \caption{The distribution of maximum (\textbf{top}) and minimum (\textbf{bottom}) thrusts over 200 randomized trajectories for three sets of $\alpha$-scaled planners: minimum snap (\textbf{left}), minimum jerk (\textbf{middle}), and minimum acceleration (\textbf{right}). In each set, we compare against an $\alpha$-scaled flatness-based planner (\textbf{blue}), an $\alpha$-scaled TOPP planner subject to velocity constraints (\textbf{orange}), and an $\alpha$-scaled TOPP planner subject to velocity and thrust constraints (\textbf{green}). Only TOPPQuad (\textbf{red}) is able to take full range of the quadrotor's inputs.}
    \label{fig:motor_speed_feasible}
\end{figure}

We measure dynamic feasibility by the percentage of trajectories that stay within motor thrust bounds. Fig. \ref{fig:motor_speed_infeasible} shows that although trajectories have constraints on velocity and acceleration, commonly used convex approximations, they often require inputs that strongly exceed allowed motor thrust bounds. In Fig. \ref{fig:motor_speed_feasible}.  we focus only on the motor thrusts of dynamically feasible planners, including TOPPQuad. While all planners generate trajectories that respect motor thrust bounds, the three baseline methods are conservative compared to the TOPPQuad trajectories, and do not always take full advantage of the quadrotor's flight capabilities.

\subsection{Time Optimality}

\begin{figure}[h]
    \centering
    \vspace{5pt}
    \includegraphics[width=\linewidth]{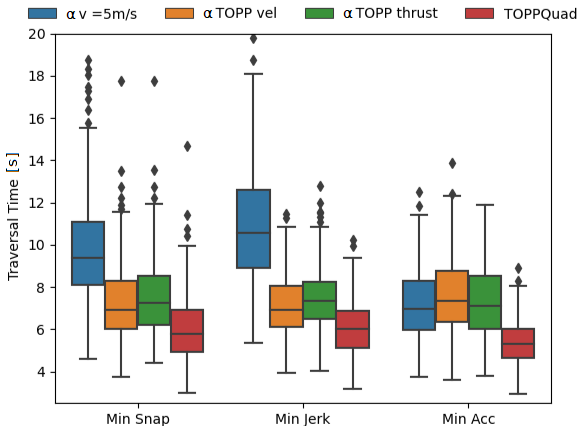}
    \caption{The distribution of traversal times over 200 randomized trajectories for three sets of $\alpha$-scaled planners: minimum snap (\textbf{left}), minimum jerk (\textbf{middle}), and minimum acceleration (\textbf{right}). In each set, we compare the traversal time for a $\alpha$-scaled flatness-based planner (\textbf{blue}), an $\alpha$-scaled TOPP planner subject to velocity constraints (\textbf{orange}), an $\alpha$-scaled TOPP planner subject to velocity and thrust constraints (\textbf{green}), and our TOPPQuad algorithm (\textbf{red}).}
    \label{fig:time_optimality}
\end{figure}

\begin{figure}[h]
    \centering
    \vspace{5pt}
    \includegraphics[width=\linewidth]{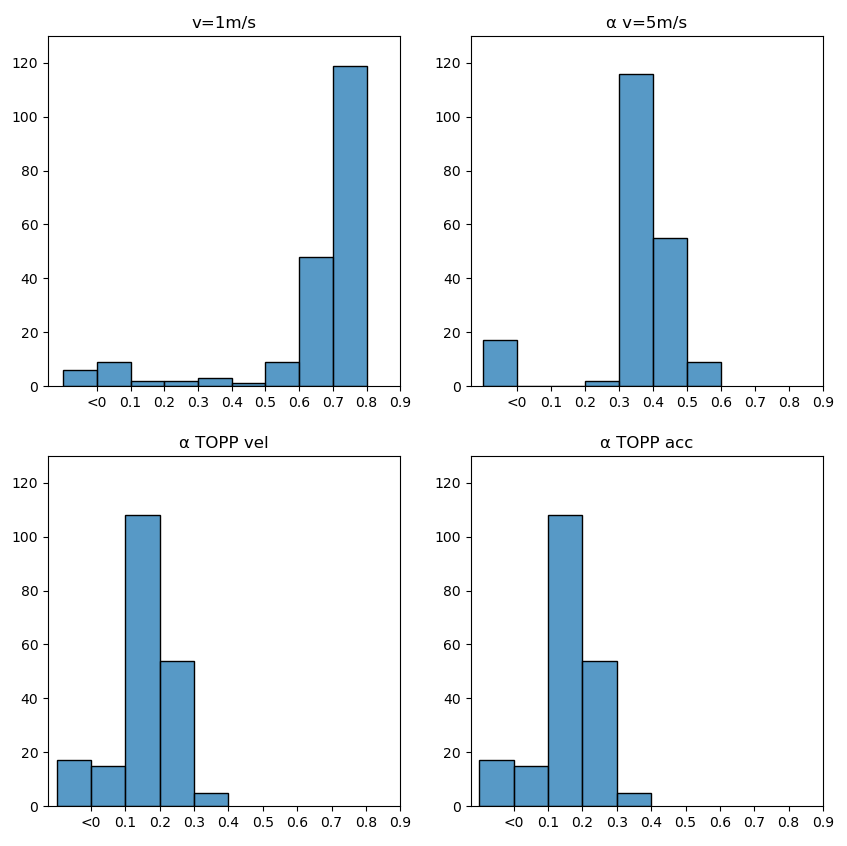}
    \caption{The distribution of traversal time improvement over the four dynamically feasible planners when refined with TOPPQuad: min snap $v=1m/s$ (\textbf{top left}), $\alpha$-scaled $v=5m/s$ min snap (\textbf{top right}), $\alpha$-scaled TOPP w/velocity constraints (\textbf{bottom left}), and $\alpha$-scaled TOPP w/acceleration constraints (\textbf{bottom right}). Values are computed against the initial $v=1m/s$ trajectory guess. Our algorithm consistently sees improvements in total traversal time, even against other time-optimal methods.}
    \label{fig:success_hist}
\end{figure}

Next, we focus on the time optimality of the dynamically feasible trajectory planners: the $\alpha$-scaled methods and TOPPQuad. Fig. \ref{fig:time_optimality} shows that our TOPPQuad method consistently generates faster trajectories than the $\alpha$-scaled methods. We see anywhere between a $2-5$ sec decrease from the unconstrained trajectories and about a $2$ sec decrease from the TOPP trajectories. This corresponds to an approximate $1.8m/s$ and $0.8m/s$ increase in respective average trajectory velocity. Note that the shorter average speed of the minimum acceleration trajectories is due to the faster average trajectory, not algorithm performance. 

In Fig. \ref{fig:success_hist}, we show the percentage time decrease between our method and the three dynamically feasible baselines for the minimum snap case, as well as the $v=1m/s$ initial guess. For the $\alpha$-scaled minimum snap planner, the majority of trajectories see a time decrease of $40\%$ or more, and even the TOPP methods see at least a $10\%$ decrease.

\subsection{Bidirectionality}

The TOPPQuad planner can be easily adapted for optimizing trajectories for bidirectional quadrotors while sidestepping the difficulties of planning for a flatness-based model \cite{Mao2023}.
Such vehicles are governed by the same set of rigid body dynamics as 'unidirectional' quadrotors, but can also exert thrust along the negative body $z$-axis of the robot. 


We explore how much efficiency can be gained with bidirectional motors. At $v_{max}=5m/s$, we see a traversal time performance on par with or only marginally faster than the unidirectional quad. This can likely be attributed to the saturation of motor thrusts and the maximum velocity constraint, preventing further time improvements.

In Fig. \ref{fig:bidirectional_comp}, we show the same trajectory computed for a time-optimal unidirectional and bidirectional quadrotor, with the maximum speed capped at $v=10m/s$. 
While the total traversal time improved by only around $0.2$s, the orientations along the path of the bidirectional trajectory appear smoother than their unidirectional counterparts. We suspect this is due to the quadrotor's ability to temporarily dip into the negative thrust regime, thus allowing for more rapid orientation change and longer saturation of maximum motor thrusts. The addition of motor thrust rate of change constraints may dampen this effect, but this initial exploration demonstrates the potential for quadrotors with bidirectional thrust in traversing time optimal trajectories.

\begin{figure}[h]
    \centering
    \vspace{5pt}
    \includegraphics[width=\linewidth]{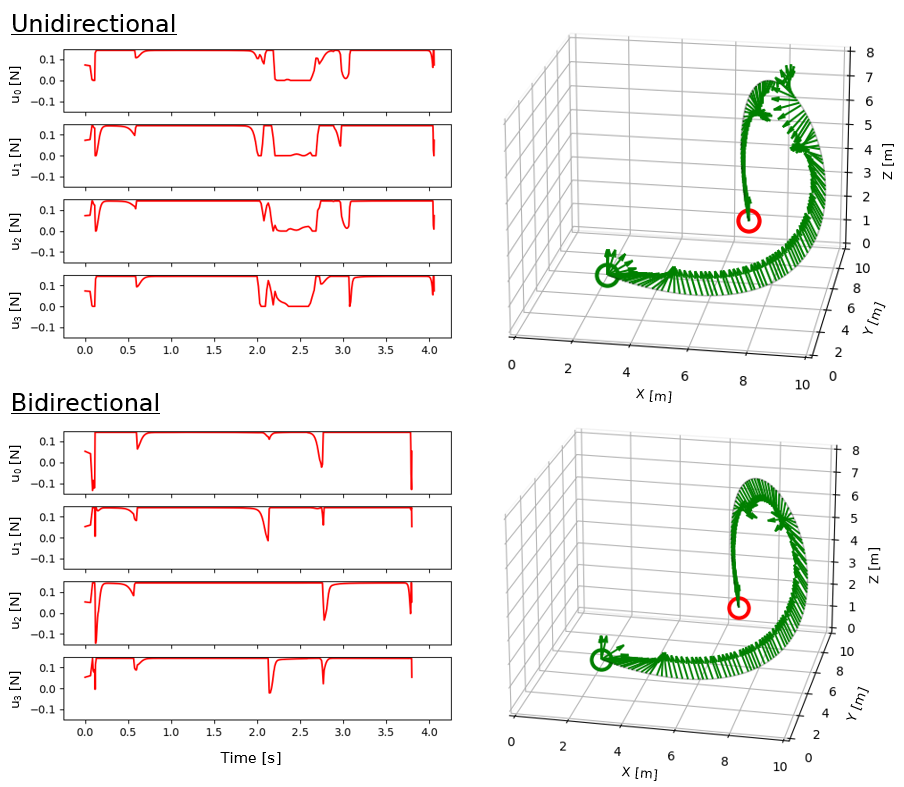}
    \caption{A sample trajectory computed with TOPPQuad for both a unidirectional (\textbf{top}) and bidirectional (\textbf{bottom}) quadrotor, and the respective motor speeds. The green vectors represent the quadrotor's $\mathbf{R}\mathbf{e}_3$ vector at each point along the trajectory.}
    \label{fig:bidirectional_comp}
\end{figure}

\subsection{TOPPQuad in the Wild}
\begin{figure}[h]
    \centering
    \includegraphics[width=\linewidth]{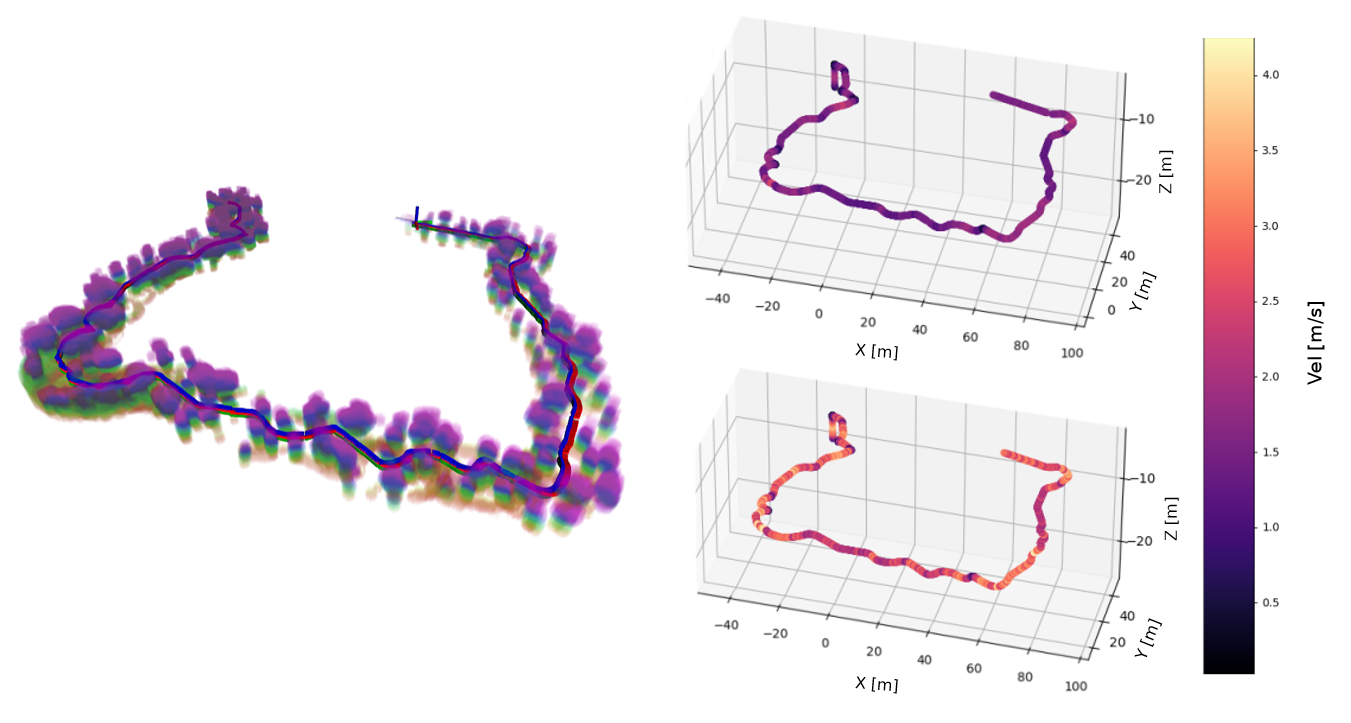}
    \caption{Obstacle-free path through a forest (\textbf{left}) . We show the quadrotor speed along the original path (\textbf{top right}) and refined TOPPQuad path (\textbf{Top left}). Maximum velocity is bounded at $v=5m/s$.}
    \label{fig:falcon_flight}
\end{figure}

Here, we demonstrate the performance of our planner on real-world flight data, when not all initial guess values are known. We apply TOPPQuad to a trajectory flown in the Wharton State Forest, NJ, USA, where a Falcon IV quadrotor autonomously navigated under the canopy of a large, tree-dense environment (\cite{9812235}, \cite{9720974}).
We recover the geometric path and the initial guess for TOPPQuad from the flight's odometry data. As we do not have estimates of motor speeds or angular acceleration, we use the values of the hover configuration ($4108$ $rad/s$, $0$ $rad/s^2$) along the full trajectory.
Fig. \ref{fig:falcon_flight} shows the increase in speed along the path. The total trajectory time decreases from $236$ $sec$ to $136$ $sec$ across a total trajectory length of $336.2$ $m$.

\section{Deployment Results}

We validate the performance of our TOPPQuad trajectories with deployment on a real-world quadrotor platform, compared against minimum snap trajectories. We use a Crazyflie 2.0 from Bitcraze, with a nominal weight of $32g$. All experiments were performed in a VICON motion capture system to collect the ground truth state information. 
Trajectories were computed offboard and commands sent to the quadrotor from a computer base-station at a frequency of 100 Hz. 


We use the $SE(3)$ Geometric Controller \cite{5717652} for trajectory tracking. PD controller gains are listed in Table \ref{tab:gains}.  Given the controller samples at a higher frequency than the  TOPPQuad trajectory discretization, we rely on an interpolation of points to determine the requisite flat outputs at each time stamp. For the positional flat outputs (position, velocity, and yaw), we fit a $5th$-order polynomial to the corresponding pair of positions, velocitues, and accelerations. For yaw, we fit a quaternion spline \cite{kim1995general} to the corresponding pair of quaternions and angular velocities. This approach naturally results in sub-time-optimal trajectory, as the interpolation will smooth away any high-frequency behavior that may occur between sampled points of the TOPPQuad trajectory. Nevertheless, the resulting trajectories demonstrate good trackability and a trajectory run-time improvement over their minimum snap counterparts.

\begin{table}[h]
\centering
\vspace{5pt}
\begin{tabular}{lllll}
           & \textbf{$K_p$} & \textbf{$K_d$} & \textbf{$K_R$} & \textbf{$K_\omega$} \\ \hline
\textbf{x} & 8            & 5.5          & 2812         & 128          \\
\textbf{y} & 8            & 5.5          & 2812         & 128          \\
\textbf{z} & 19           & 8.7          & 163          & 73           \\ \hline
\end{tabular}
\caption{SE(3) Controller Gains}
\label{tab:gains}
\vspace{-15pt}
\end{table}



The following four trajectories are shown: a line (Fig. \ref{fig:exp_line}), an L-shaped curve (Fig. \ref{fig:exp_cube}), an X-shaped curve in 3D space (Fig. \ref{fig:exp_curve}), and a Lissajous curve (Fig. \ref{fig:TOPPQUad}, \ref{fig:exp_liss}).  For the first three trajectories, we compare the trackability of two minimum snap (MS) trajectories, with heuristic velocities of $1m/s$ and $2m/s$, against a TOPPQuad trajectory with a maximum velocity constraint of $2m/s$. We show a consistent decrease in execution time between the $1m/s$ MS and TOPPQuad trajectories and improved trackability between the $2m/s$ MS (which all crashed) and TOPPQuad trajectories. In addition, we compare our planner with a more demanding Lissajous trajectory. We show that we are able to track the TOPPQuad-refined Lissajous curve while the original trajectory fails. In all experiments, we consider a trajectory executed once the state of the robot settles to the terminal state of the trajectory.

\begin{figure}[h]
    \centering
    \vspace{5pt}
    \includegraphics[width=\linewidth]{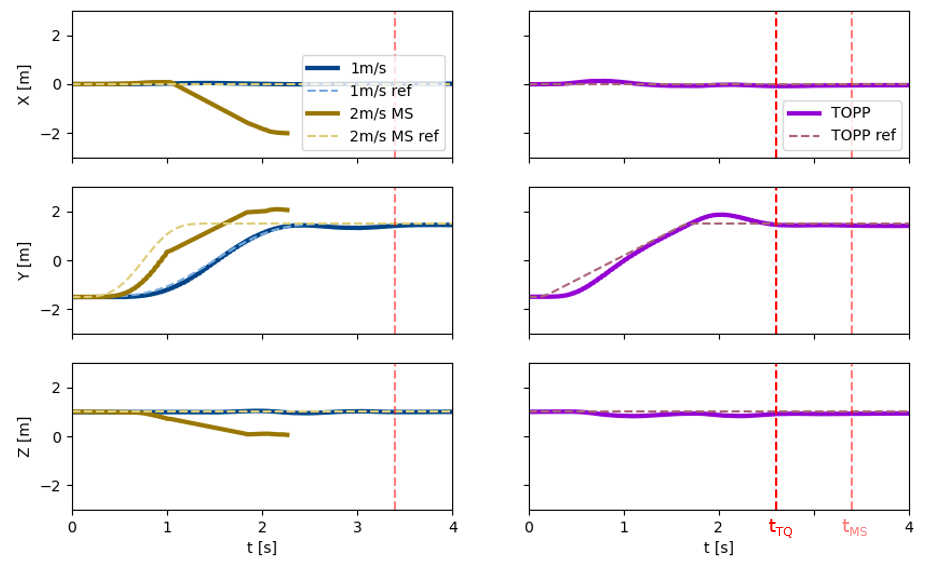}
    \caption{\textbf{Line}. \textit{Left:} Plots of the  tracked (blue, purple) and reference (orange, yellow) $1m/s$ MS and $2m/s$ MS trajectories in X, Y, Z. \textit{Right:} Plots of tracked (purple) and reference (blue) TOPPQuad trajectory. With the controller in the loop, the TOPPQuad trajectory was executed in $t_{TQ} = 2.6s$, the $1m/s$ MS trajectory executed in $t_{MS} = 3.4s$, and the $2m/s$ MS trajectory crashed.}
    \label{fig:exp_line}
\end{figure}

\begin{figure}[h]
    \centering
    \includegraphics[width=\linewidth]{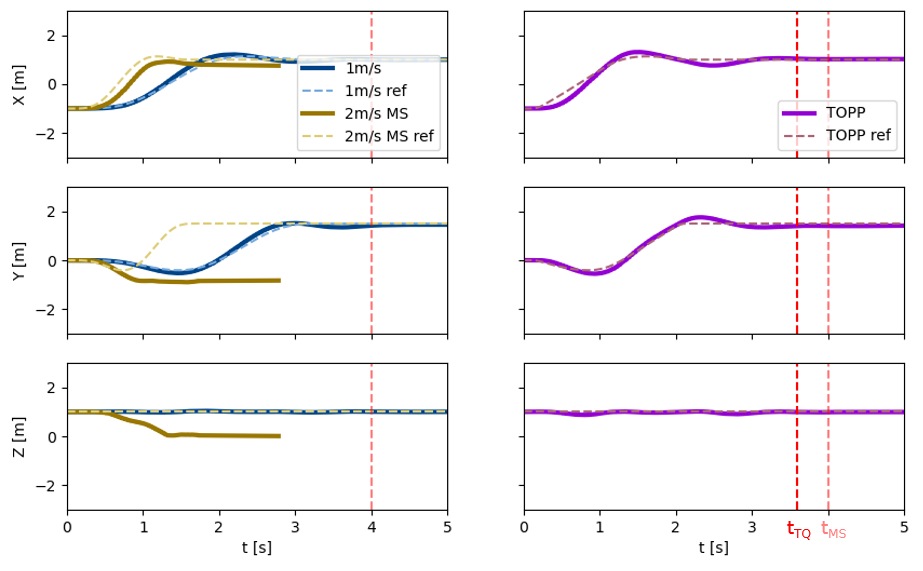}
    \caption{\textbf{L-Curve}. \textit{Left:} Plots of the tracked (blue, purple) and reference (orange, yellow) $1m/s$ MS and $2m/s$ MS trajectories in X, Y, Z. \textit{Right:} Plots of tracked (purple) and reference (blue) TOPPQuad trajectory. With the controller in the loop, the TOPPQuad trajectory was executed in $t_{TQ} = 3.6s$, the $1m/s$ MS trajectory executed in $t_{MS} = 4s$, and the $2m/s$ MS trajectory crashed.}
    \label{fig:exp_cube}
\end{figure}

\begin{figure}[h]
    \centering
    \vspace{5pt}
    \includegraphics[width=\linewidth]{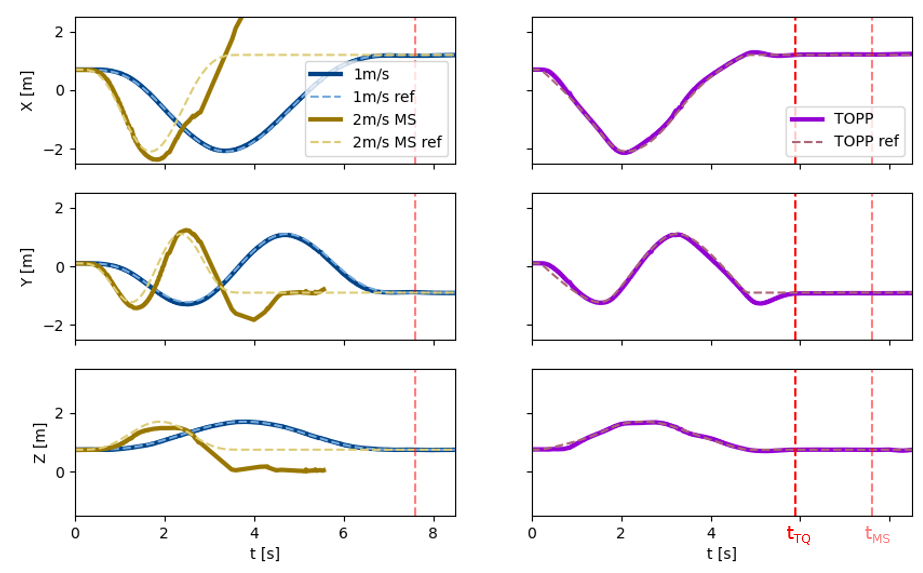}
    \caption{\textbf{X-Curve}. \textit{Left:} Plots of the  tracked (blue, purple) and reference (orange, yellow) $1m/s$ MS and $2m/s$ MS trajectories in X, Y, Z. \textit{Right:} Plots of tracked (purple) and reference (blue) TOPPQuad trajectory. With the controller in the loop, the TOPPQuad trajectory was executed in $t_{TQ} = 5.9s$, the $1m/s$ MS trajectory executed in $t_{MS} = 7.6s$, and the $2m/s$ MS trajectory crashed.}
    \label{fig:exp_curve}
\end{figure}

\begin{figure}[h]
    \centering
    \includegraphics[width=\linewidth]{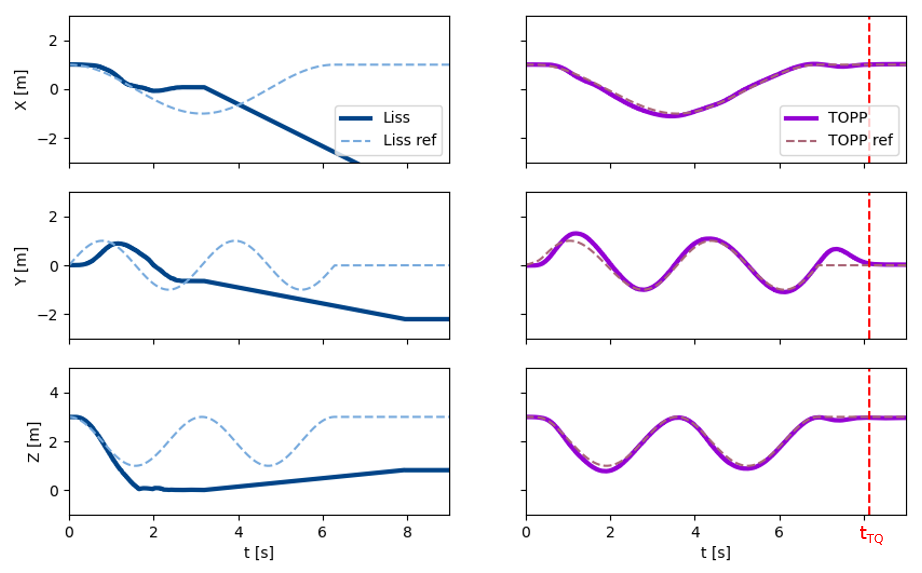}
    \caption{\textbf{Lissajous Curve}. \textit{Left:} Plots of the  tracked (blue) and reference (orange) Lissajous trajectory in X, Y, Z. \textit{Right:} Plots of tracked (purple) and reference (blue) TOPPQuad trajectory. With the controller in the loop, the TOPPQuad trajectory was executed in $t_{TQ} = 9.1s$, while the Lissajous trajectory crashed.}
    \label{fig:exp_liss}
\end{figure}

\section{Conclusion}

We present TOPPQuad, an optimization algorithm for finding time-optimal quadrotor trajectories. We explicitly enforce state and input constraints such as motor thrusts
instead of relying on convex relaxations, which we demonstrate are not always sufficient. The key to our algorithm is an optimization of the total trajectory time given the full dynamics of the quadrotor. We show the ability to refine trajectories that come from a variety of commonly used differential flatness-based planners, expand motor speed bounds to automatically generate trajectories for bidirectional quadrotors, and demonstrate our planner's performance on real world data. This offers a complimentary approach to existing time-optimal quadrotor planning methods.

However, our algorithm still has room for improvement time. Firstly, the computation time takes on the order of half a minute. Secondly, by virtue of solving a non-convex problem, our planner is not complete. However, fall-backs such as $\alpha$-scaling offer completeness, although without guarantees of time-optimality. Finally, our model does not account for drag or motor dynamics.
Future work will focus on examining implementations for improving algorithm computation time and will focus on a higher fidelity model by incorporating representations of air resistance and motor dynamics into the optimization framework.

\bibliographystyle{IEEEtran}
\bibliography{references}

\end{document}